\pdfoutput=1

\documentclass[11pt]{article}

\usepackage[final]{acl}

\usepackage{times}
\usepackage{latexsym}

\usepackage[T1]{fontenc}

\usepackage[utf8]{inputenc}

\usepackage{microtype}

\usepackage{inconsolata}

\usepackage{graphicx}

\usepackage{multicol}
\usepackage{multirow}
\usepackage{booktabs}
\usepackage{makecell}
\usepackage{colortbl}
\usepackage{xcolor}
\usepackage{amssymb}
\usepackage{amsmath}
\usepackage{diagbox}
\usepackage[ruled,linesnumbered]{algorithm2e}

\title{KCS: Diversify Multi-hop Question Generation with Knowledge Composition Sampling}
  
\author{Yangfan Wang\textsuperscript{1}, Jie Liu\textsuperscript{1,3}, Chen Tang\textsuperscript{2}, Lian Yan\textsuperscript{1}, Jingchi Jiang\textsuperscript{1,3}\footnotemark[1] \\
\textsuperscript{1}Harbin Institute of Technology, \textsuperscript{2}MemTensor (Shanghai) Technology Co., Ltd. \\ \textsuperscript{3}National Key Laboratory of Smart Farm Technologies and Systems \\
\texttt{\{yf.wang,23b903008\}@stu.hit.edu.cn}, \texttt{travistang@foxmail.com} \\
\texttt{\{jieliu,jiangjingchi\}@hit.edu.cn}
}

\begin{document}
\maketitle

\renewcommand{\thefootnote}{\fnsymbol{footnote}} 
\footnotetext[1]{Corresponding author.} 
\renewcommand*{\thefootnote}{\arabic{footnote}}

\begin{abstract}

Multi-hop question answering faces substantial challenges due to data sparsity, which increases the likelihood of language models learning spurious patterns. To address this issue, prior research has focused on diversifying question generation through content planning and varied expression. However, these approaches often emphasize generating simple questions and neglect the integration of essential knowledge, such as relevant sentences within documents. This paper introduces the \textbf{Knowledge Composition Sampling (KCS)}, an innovative framework designed to expand the diversity of generated multi-hop questions by sampling varied knowledge compositions within a given context. KCS models the knowledge composition selection as a sentence-level conditional prediction task and utilizes a probabilistic contrastive loss to predict the next most relevant piece of knowledge. During inference, we employ a stochastic decoding strategy to effectively balance accuracy and diversity. Compared to competitive baselines, our KCS improves the overall accuracy of knowledge composition selection by 3.9\%, and its application for data augmentation yields improvements on HotpotQA and 2WikiMultihopQA datasets. Our code is available at: \url{https://github.com/yangfanww/kcs}.

\end{abstract}

\begin{figure}[!t]
  \centering
  \includegraphics[width=\columnwidth]{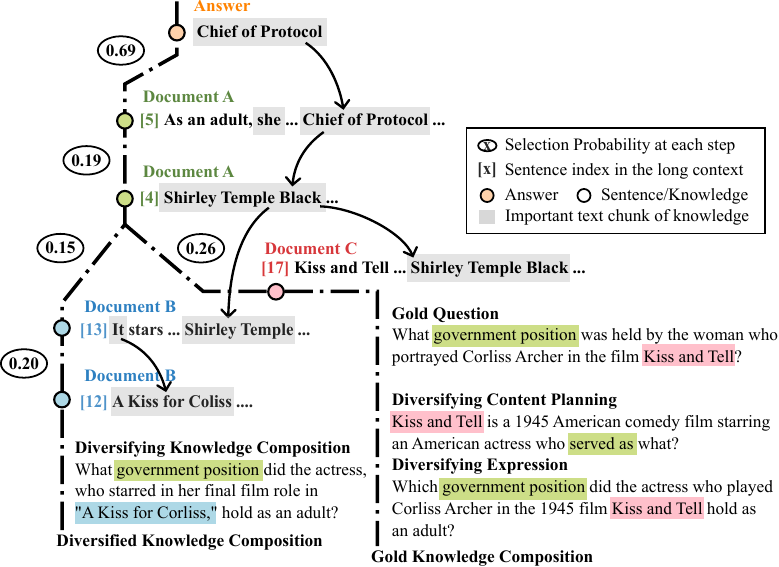}
  \caption{This figure illustrates the diversity of contextual knowledge through a tree structure. Starting from an answer (root), each subsequent knowledge piece is selected from the long context based on a probability (edge). Each knowledge composition is depicted as a branch. Unlike prior methods that consistently select the same knowledge composition as the gold standard (right), KCS tends to select varied branches to enhance the diversity of generated multi-hop questions (left).
  }
  \label{fig:example}
\end{figure}

\section{Introduction}

Multi-hop Question Answering (MHQA) presents unique challenges in natural language processing \citep{panda-etal-2024-holmes}, requiring the selection and integration of multiple knowledge pieces to accurately answer complex questions. Despite advancements facilitated by high-quality MHQA datasets \citep{yang-etal-2018-hotpotqa,ho-etal-2020-constructing}, data sparsity remains a significant issue, increasing the risk for language models learning spurious patterns and compromising robustness and generalization. Question Generation (QG) has been proposed to augment QA datasets \citep{guo-etal-2022-dsm}. Recent efforts to diversify QG focus on enhancing question diversity through content planning and expression, aiming to capture the one-to-many nature of QG tasks \citep{gou-etal-2023-diversify,narayan-etal-2022-well,deschamps-etal-2021-generating,cao-wang-2021-controllable,DBLP:conf/iclr/HoltzmanBDFC20,shen2019mixture,cho-etal-2019-mixture}. These methods often concentrate on the generation of simple questions and fail to introduce external knowledge, such as relevant sentences from documents, to generate more complex and diverse questions. In reality, without incorporating external knowledge and relying solely on simple textual methods, the diversity of questions doesn't truly improve. Identifying important sentences that encapsulate contextually relevant knowledge is essential for facilitating the model's understanding of the underlying logic in the data, thus generating questions deemed semantically diverse and meaningful by humans  \citep{du-cardie-2017-identifying}. Given the inherent complexity of MHQA (a task that integrates multiple pieces of knowledge), this step is crucial for diversifying multi-hop QG. We assert that data sparsity stems from the \textit{underutilization of contextual knowledge}. As illustrated in Figure~\ref{fig:example}, selecting a different set of slightly varied knowledge pieces (termed a knowledge composition) can lead a QG model to generate a significantly distinct multi-hop question at the knowledge level, even when the answer remains unchanged. This observation suggests that contextual knowledge is not fully utilized in existing MHQA datasets. The detialed example is shown in Appendix~\ref{appendix:example}.

To address these challenges, we propose the \textbf{Knowledge Composition Sampling (KCS)}, a novel framework based on knowledge diversity that facilitates the utilization of contextual knowledge by sampling varied knowledge compositions within a given context, thus enhancing multi-hop question diversity. Our framework contains three main components: (1) knowledge composition selection to accurately select knowledge compositions from context; (2) diversifying knowledge composition to efficiently sample accurate and diverse knowledge compositions; and (3) multi-hop question generation using a vanilla model to generate multi-hop questions from the given answer and sampled knowledge compositions. 

However, as shown in Figure~\ref{fig:_example}, arbitrary knowledge compositions pose the risk of \textit{degeneration}\footnote{\citet{DBLP:conf/iclr/HoltzmanBDFC20} define ``degeneration'' as the production of automatically generated text that is generic and repetitive. \citet{narayan-etal-2022-well} define it as text that is unfaithful or inconsistent with the input. In our context, ``degeneration'' refers to irrelevant knowledge leading to simpler or inconsistent multi-hop questions.}. To mitigate this risk, we frame knowledge composition selection as a sentence-level conditional prediction problem and utilize a probabilistic contrastive loss to learn the potential knowledge coherence. During training, the selection model maximizes mutual information between the latent prediction representation of the current timestep and the latent representation of the next timestep, while minimizing mutual information with other latent representations within the context. To balance the accuracy and diversity, we employ a stochastic decoding strategy that truncates the unreliable tail of the probability distribution and samples the next knowledge piece from a dynamic nucleus. As shown in Figure~\ref{fig:example}, the selection of each subsequent sentence is conditional on the answer, context, and previously selected sentences. By modeling the conditional probabilities of each timestep and employing stochastic sampling, KCS effectively obtain diversified and accurate knowledge compositions, thereby producing diverse and high-quality multi-hop questions. In summary, unlike traditional methods that rely on structured graphs or textual methods on fixed knowledge compositions, our KCS leverages unstructured text to discern the potential knowledge coherence, enhancing flexibility and scalability. Our contributions are as follows:

\begin{itemize}
    \setlength{\itemsep}{0pt}
    \item We propose the KCS framework designed to expand the diversity of generated multi-hop questions by sampling varied knowledge compositions within a given context;
    \item We introduce a novel sentence-level conditional prediction task and a probabilistic contrastive loss to discern potential knowledge coherence, and verify the effectiveness of a stochastic decoding strategy;
    \item Experiments on HotpotQA and 2WikiMultihopQA demonstrate that KCS improves the overall accuracy of knowledge composition selection by 3.9\%, and its use for data augmentation achieves consistent improvements of downstream performance.
\end{itemize}

\begin{figure*}[t]
  \centering
  \includegraphics[width=0.95\linewidth]{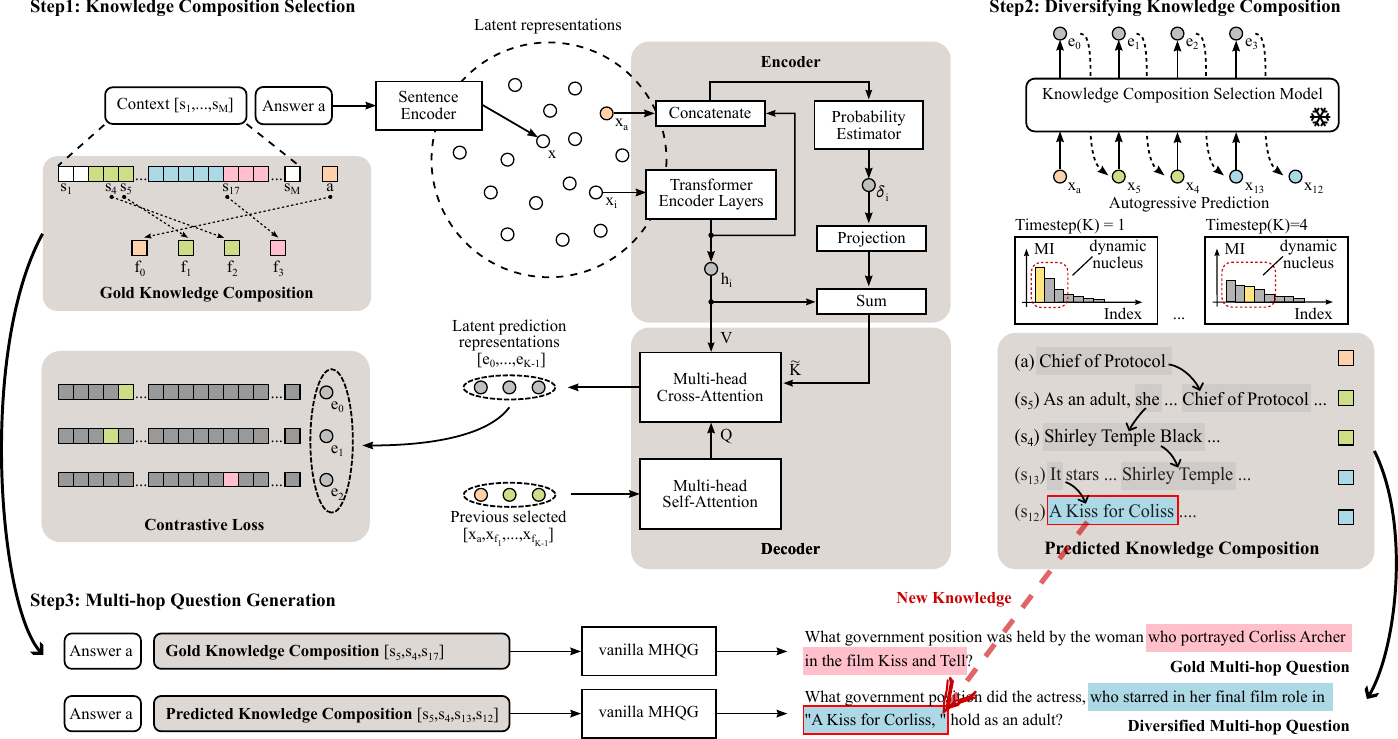}
  \caption {Illustration of our KCS framework.}
  \label{fig:framework}
\end{figure*}

\section{Related Works}

\paragraph{Important Sentence Selection} The initial step in QG task involves identifying sentences within the context that are question-worthy, i.e., sentences that humans consider valuable for generating questions. Previous research \citet{du-cardie-2017-identifying} predominantly formalizes this task as sentence classification, often focusing on simple question generation. In our study, we introduce sentence-level sequence prediction to enhance the selection of knowledge pieces for multi-hop question generation (MHQG).

\paragraph{Multi-hop Question Generation} Previous research \citep{kumar2019difficulty, pan-etal-2020-semantic, fei-etal-2022-cqg, hwang-etal-2024-explainable} frequently relies on pre-constructed knowledge graphs, entity graphs, or document graphs to enable controllable MHQG. This dependency often results in errors and increased costs associated with entity extraction and graph construction. Alternative approaches leverage in-context learning with large language models \citep{lin-etal-2024-prompting} or employ supervised fine-tuning of pretrained language models \citep{murakhovska-etal-2022-mixqg, xia2023improving} to achieve consistent MHQG. Our framework samples varied knowledge compositions from the given context for diversifying MHQG, allowing for seamless integration with existing MHQG methods. In our study, we fine-tune a vanilla model proposed by \citep{murakhovska-etal-2022-mixqg} for MHQG.

\paragraph{Diversifying Question Generation} Recent works in diversifying QG have aimed to enhance the diversity of generated questions across two dimensions: content planning and expression. For content planning, methods such as \citep{cho-etal-2019-mixture} sample diverse token-level content, while others \citep{narayan-etal-2022-well, DBLP:conf/iclr/HoltzmanBDFC20} employ advanced decoding strategies at the token level. To achieve expression diversity, studies such as \citep{deschamps-etal-2021-generating, gou-etal-2023-diversify} utilize external knowledge to generate multiple expressions of the same question. However, these methods primarily focus on simple questions and assume that important sentences (knowledge compositions) in context are pre-identified and fixed. In contrast, our work addresses the entire QG process and expands the diversity of generated multi-hop questions through sampling diverse knowledge-level compositions.

\section{Method}

\subsection{Overview}
\label{sec:overview}

We formally define the problem of diversifying multi-hop QG as follows. Given a long context $D$, which consists of a set of documents, and an answer $a=[a_1, \dots, a_{T_a}]$, the objective is to generate a set of diverse multi-hop questions $Q=\{q_1, \dots, q_{N_q}\}$, conditioned on both $D$ and $a$. Here, $T_a$ denotes the number of tokens in $a$, and $N_q$ represents the number of generated questions. Each document within the context is composed of multiple sentences, such that the context $D=[s_1, \dots, s_M]$ contains $M$ sentences in total. We distinguish between the long context $D$, which encapsulates extensive knowledge, and a knowledge composition $c=[f_1, \dots, f_K]$, a subset of $D$ containing $K$ question-worthy sentences. Previous approaches term $c$ as ``context'' and typically model the conditional probability $p(q|c,a)$, with the assumption that $c$ is pre-identified and fixed. In contrast, we address the entire pipeline of diversifying multi-hop QG without such assumptions, modeling the problem as follows:

\begin{equation}
\hspace{-0.5cm}
\begin{aligned}
  \label{eq:problem} 
  p(q|D, a) = \mathbb{E}_{c}[p(q|c,a)\times p(c|D,a)]
\end{aligned}
\hspace{-0.5cm}
\end{equation}

The overall architecture of our KCS framework is illustrated in Figure~\ref{fig:framework}. The knowledge composition selection component learns a Transformer-based sentence-level sequence prediction model to accurately select a knowledge composition $c$ for the given answer $a$ and long context $D$. The diversifying knowledge composition component employs a stochastic decoding strategy to sample diverse knowledge compositions $C=\{c_1,\dots,c_{N_q}\}$ from $D$. The multi-hop question generation component utilizes a vanilla MHQG model to generate a multi-hop question $q$ conditioned on each sampled knowledge composition $c\in C$ and the given answer $a$. The algorithm of diversifying multi-hop QG is shown in Appendix~\ref{appendix:algorithm}.

\subsection{Knowledge Composition Selection}

We employ a hierarchical neural network architecture for sentence-level sequence prediction to select accurate knowledge compositions. As depicted in Step1 of Figure~\ref{fig:framework}, the hierarchical neural network architecture includes a sentence encoder $\mathcal{M}_{\rm enc}$, a sentence-level sequence model $\mathcal{M}_{\rm seq}$, and two objectives $\mathcal{L}_{\rm cls}$ and $\mathcal{L}_{\rm seq}$, which are aligned with human intuitions. Formally, the classification objective $\mathcal{L}_{\rm cls}$ involves assigning the correct label $z\in [0,1]$ to each sentence $s\in D$, resulting in $Z=[z_1,\dots,z_M]$. The sequence prediction objective $\mathcal{L}_{\rm seq}$ involves predicting the correct knowledge composition $c$ auto-regressively, with $f_0=a$ for convenience:

\begin{equation}
\hspace{-0.5cm}
\begin{aligned}
  \label{eq:sequence}
  p(c|D, a) & = p([f_1,\dots,f_K]|D,a) \\
  & = \prod\limits_{k=1}^{K}p(f_k|D,Z,f_{0:k-1})
\end{aligned}
\hspace{-0.5cm}
\end{equation}

First, the BERT-based $\mathcal{M}_{\rm enc}$ \citep{devlin2019bert} maps each sentence $s \in D \cup \{a\}$ to a latent representation $x = \mathcal{M}_{\rm enc}(s)$. For the long context $D$, we obtain $X=[x_1, \dots, x_M]\in\mathbb{R}^{M\times d}$, where $d$ is the hidden state dimension. Next, the sentence-level Transformer-based model $\mathcal{M}_{\rm seq}$ \citep{vaswani2017attention} extends the Transformer's encoder to produce knowledge classification probabilities $Z$, which are then infused from its encoder into its decoder.

\begin{equation}
\hspace{-0.5cm}
\begin{aligned}
  \label{eq:encoding}
  H & = {\rm EncoderLayers}(X) \\
  z_i & = {\rm Softmax}({\rm Linear}(h_i;x_a)) \\
  e_{k-1} & = {\rm DecoderLayers}(x_{f_{\le k-1}},H,Z)
\end{aligned}
\hspace{-0.5cm}
\end{equation}

Specifically, the encoder of $\mathcal{M}_{\rm seq}$ encodes the latent representations $X$ to hidden states $H=[h_1,\dots,h_M]\in\mathbb{R}^{M\times d}$. Each $h_i$ is then concatenated with the latent answer representation $x_a$, and a linear network with softmax serves as the probability estimator to obtain the knowledge classification probability $p(z_i|x_a,h_i)$. We input hidden states $H$ to the multi-head cross-attention layer of the decoder as the value state but modify the key state as $\tilde{K}= H+\delta W^{\delta}$ where $\delta_i=[1-z_i, z_i]$, using a linear projection with parameter $W^{\delta}\in\mathbb{R}^{2\times d}$. The decoder consumes the latent representations of previously selected sentences $f_{\le k-1}\subset D$ to produce a latent prediction representation $e_{k-1}$ and predict the conditional selection probability $p(s|e_{k-1})$.

Inspired by the noise-contrastive estimation in \citep{oord2018representation}, we discard low-level information and noise and introduce a probabilistic contrastive loss for next step prediction to learn the potential knowledge coherence. Formally, given the long context $D=[s_1, \dots, s_M]$, we treat the next $f_k$ as the sole positive sample from $p(s|e_{k-1})$ and others as negative samples from the distribution $p(s)$ at the $(k-1)^{\rm th}$ timestep. We maximize mutual information $\rm MI$ between the latent prediction representation $e_{k-1}$ of the current timestep and the latent representation $x_{f_k}$ of the next timestep, while minimizing mutual information with other latent representations in the context, optimizing the probabilistic contrastive loss $\mathcal{L}_{\rm seq}$:

\begin{equation}
\hspace{-0.5cm}
\begin{aligned}
  \label{eq:loss}
  \mathcal{L}_{\rm seq} & = -\mathbb{E}_c\left[\log{\frac{{\rm MI}(x_{f_k}, e_{k-1})}{\sum\limits_{s\in D,s\ne f_k}{\rm MI}(x_s, e_{k-1})}}\right] \\
\end{aligned}
\hspace{-0.5cm}
\end{equation}

where $\rm MI$ is a mutual information function. The final loss is $\mathcal{L}= \mathcal{L}_{\rm cls} + \lambda\mathcal{L}_{\rm seq}$, where $\mathcal{L}_{\rm cls}$ is the classification loss, and $\lambda$ is a hyper-parameter.

\subsection{Diversifying Knowledge Composition}

\begin{figure}[!h]
  \centering
  \includegraphics[width=0.9\columnwidth]{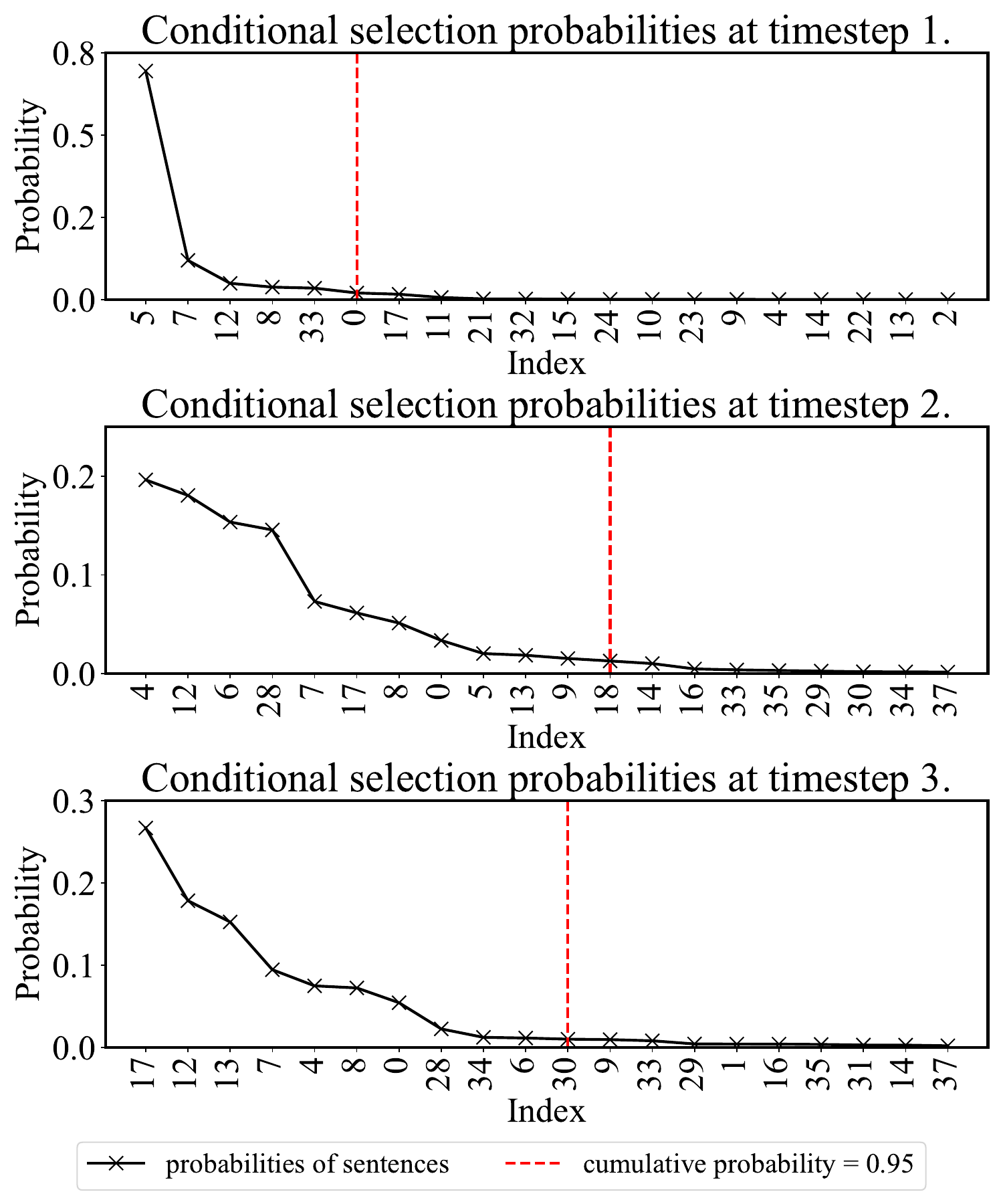}
  \caption{Conditional probabilities for knowledge selection at each timestep, with sentence indexes in descending order of probabilities. The \textcolor{red}{red dashed line} indicates the cutoff of $p=0.95$. We sequentially select the sentence with the highest probability as the next knowledge for greedy sampling, i.e., \texttt{[5,4,17]} form the knowledge composition.}
  \label{fig:nucleus}
\end{figure}

While maximization-based decoding strategies are effective in selecting accurate knowledge compositions, they often lack diversity. As shown in Figure~\ref{fig:nucleus}, the conditional probability distribution $p(s|e_{k-1})$ of our knowledge composition selection model exhibits an unreliable tail that requires truncation during generation. We introduce a stochastic decoding strategy to efficiently sample accurate and diverse knowledge compositions. Figure~\ref{fig:nucleus} aligns with the example in Figure~\ref{fig:example} and demonstrates the effectiveness of this stochastic decoding strategy.

Inspired by \citep{DBLP:conf/iclr/HoltzmanBDFC20}, we employ a nucleus sampling to the hierarchical neural network, using the shape of the probability distribution to determine the set of sentences to be sampled at each timestep. We truncate the unreliable tail of the conditional probability distribution at each timestep, and then sample the next sentence from a dynamic nucleus of sentences that contains the majority of the probability mass. Formally, given the conditional probability distribution $p(s|e_{k-1})$, we define its top-$p$ nucleus $D^{(p)}\subset D$ as the smallest set such that $\sum_{s\in D^{(p)}} p(s|e_{k-1}) \ge p$. The original distribution is then rescaled to form a new distribution $p_{\rm new}$ from which the next sentence is sampled.

\subsection{Multi-hop Question Generation}
\label{sec:mhqg}

As illustrated in Figure~\ref{fig:framework}, our KCS framework separates the diversification process from the generation process. The one-to-many process is accomplished at the knowledge level through the previous two components, resulting in accurate and diverse knowledge compositions. Given each identified knowledge composition $c$ and the answer $a$, the multi-hop question generation component aims to generate a consistent multi-hop question $q$. For the same answer and varied knowledge compositions $C=[c_1,\dots,c_{N_q}]$, we generate a set of diverse multi-hop questions $Q=\{q_1,\dots,q_{N_q}\}$.

Specifically, we fine-tune a vanilla MixQG-base model \citep{murakhovska-etal-2022-mixqg} on the training data, and employ the standard CrossEntropy loss to generate consistent multi-hop questions. It is noteworthy that although we use the MixQG-base model, many advanced multi-hop question generation methods can be applied to improve the consistency of the generated multi-hop questions.

\section{Experiments}

\subsection{Experimental Settings}

\paragraph{Datasets} We evaluate our method on two popular benchmark MHQA datasets: HotpotQA \citep{yang-etal-2018-hotpotqa} and 2WikiMultihopQA \citep{ho-etal-2020-constructing}. For knowledge composition selection, due to the inaccessibility of the test set for HotpotQA and the absence of supporting facts in the test set of 2WikiMultihopQA, we designate the original development set as the new test set and randomly extract 500 samples from the training set to serve as the new development set. Samples with answers of ``yes'' or ``no'' are excluded, as they do not contribute to the selection of subsequent knowledge. Detailed statistics of these datasets are provided in Appendix~\ref{appendix:statistics}. For diversifying question generation, since the evaluation datasets do not contain all possible valid questions for an answer, we indirectly evaluate the quality of generated questions through performance on downstream tasks, as a positive exploration. We randomly extract 5000 samples from the training set to serve as the original training set for data augmentation and further filter 200 samples from the new test set where each large language model (LLM) achieves the lowest Recall score to construct a test set to exclude the effect of the LLMs' base capabilities.

\paragraph{Metrics} For knowledge composition selection, we employ Precision (P), Recall (R), and F1-Score (F1) across different lengths of knowledge composition ($K=2,3$) as automatic metrics to assess the accuracy of the selected knowledge compositions. For diversifying question generation, we use Exact Match (EM), Precision (P), Recall (R), and F1-Score (F1) to evaluate the correctness of the predicted answers in the downstream MHQA task. Additionally, we compute BERTScore (BSc) and Human Evaluation Score (HSc) for semantic evaluation. Following \citep{gou-etal-2023-diversify}, we also employ Pairwise-BLEU to measure the diversity by averaging sentence-level metrics of pairs within generated $N_q$ questions, and LLM-based metrics to measure the diversity and consistency.

\begin{table*}[ht]
  \centering
  \scalebox{0.7}{
  \begin{tabular}{l|l|ccc|ccc|ccc|ccc}
  \toprule
  \multirow{2}*{\textbf{Method}} & \multirow{2}*{\textbf{Base Model}} & \multicolumn{6}{c}{\textsc{\textbf{HotpotQA}}} & \multicolumn{6}{c}{\textsc{\textbf{2WikiMultihopQA}}} \\
  & & \textit{P@2} & \textit{R@2} & \textit{F1@2} & \textit{P@3} & \textit{R@3} & \textit{F1@3} & \textit{P@2} & \textit{R@2} & \textit{F1@2} & \textit{P@3} & \textit{R@3} & \textit{F1@3} \\
  \midrule
  \multirow{7}*{\textsc{Retrieval}} & \textit{bge-small} & 48.83 & 41.49 & 44.33 & 38.19 & 48.82 & 42.22 & 40.04 & 35.50 & 37.02 & 30.69 & 41.01 & 34.53 \\
  & \textit{bge-large-v1.5} & 53.56 & 45.79 & 48.66 & 41.27 & 52.78 & 45.63 & 44.02 & 39.12 & 40.75 & 33.47 & 44.73 & 37.66 \\
  & \textit{\quad$+$step by step} & 53.78 & 46.03 & 48.90 & 42.52 & 54.29 & 46.98 & 50.91 & 43.58 & 46.03 & 40.67 & 52.24 & 44.80 \\
  & \textit{bge-m3} & 49.29 & 41.97 & 44.68 & 37.98 & 48.35 & 41.90 & 41.06 & 36.53 & 38.04 & 30.55 & 41.02 & 34.47 \\
  & \textit{text-ada-002} & 47.49 & 40.19 & 42.88 & 37.05 & 46.91 & 40.76 & 42.29 & 37.41 & 39.03 & 31.65 & 42.14 & 35.54 \\
  & \textit{BM25} & 54.44 & 46.70 & 49.57 & 40.79 & 52.34 & 45.18 & 48.10 & 43.42 & 44.98 & 35.10 & 47.38 & 39.71 \\
  & \textit{\quad$+$step by step} & 54.08 & 46.60 & 49.37 & 40.75 & 52.44 & 45.20 & 53.90 & 47.96 & 49.94 & 41.29 & 54.44 & 46.12 \\
  \midrule
  \multirow{3}*{\textsc{CLS}} & \textit{BERT-base} & 62.70 & 54.05 & 57.26 & 48.62 & 62.42 & 53.87 & \underline{79.90} & \underline{69.68} & \underline{73.09} & \underline{65.06} & \underline{82.55} & \underline{71.20} \\
  & \textit{BERT-large} & 58.00 & 49.52 & 52.64 & 44.83 & 59.69 & 49.30 & 73.25 & 64.25 & 67.25 & 61.50 & 79.00 & 67.74 \\
  & \textit{RoBERTa-large} & 23.00 & 19.52 & 20.83 & 20.33 & 25.93 & 22.47 & 44.75 & 36.88 & 39.50 & 37.83 & 46.25 & 40.60 \\
  \midrule
  \multirow{2}*{\makecell[l]{\textsc{Sentence} \\ \textsc{Graph}}} & \textit{CommonEntity} & 46.39 & 40.03 & 42.40 & 33.11 & 42.74 & 36.79 & 51.20 & 45.24 & 47.23 & 37.09 & 48.68 & 41.33 \\
  & \textit{Similarity} & 40.95 & 35.40 & 37.46 & 27.39 & 35.50 & 30.50 & 41.62 & 37.87 & 39.12 & 27.88 & 38.07 & 31.74 \\
  \midrule
  \multirow{6}*{\textsc{LLM}} & \textit{Qwen2.5-14B} & 18.10 & 15.18 & 16.26 & 14.77 & 18.49 & 16.16 & 19.24 & 16.09 & 17.14 & 16.68 & 20.84 & 18.11 \\
  & \textit{Qwen2.5-7B} & 19.26 & 16.12 & 17.28 & 15.90 & 19.92 & 17.40 & 25.05 & 21.58 & 22.74 & 19.61 & 25.20 & 21.61 \\
  & \textit{Llama3.1-8B} & 7.21 & 5.95 & 6.41 & 6.81 & 8.45 & 7.42 & 10.96 & 8.84 & 9.55 & 10.61 & 12.85 & 11.33 \\
  & \textit{Llama3.2-3B} & 9.62 & 8.08 & 8.65 & 8.11 & 10.18 & 8.88 & 13.73 & 10.91 & 11.85 & 11.71 & 14.00 & 12.42 \\
  & \textit{GPT-4} & 52.10 & 43.73 & 46.83 & 45.10 & 56.76 & 49.47 & 49.74 & 43.59 & 45.64 & 40.10 & 52.56 & 44.65 \\
  & \textit{DeepSeek-V3} & 49.93 & 41.66 & 44.71 & 43.08 & 53.95 & 47.14 & 45.77 & 40.15 & 42.02 & 37.96 & 50.09 & 42.43 \\
  \midrule
  \multirow{2}*{\textsc{BASE}} & \textit{Random} & 6.10 & 5.02 & 5.41 & 6.13 & 7.60 & 6.67 & 8.60 & 6.95 & 7.50 & 9.01 & 10.96 & 9.64 \\
  & \textit{MAX} & \textit{100.00} & \textit{87.01} & \textit{91.85} & \textit{78.00} & \textit{97.52} & \textit{85.33} & \textit{100.00} & \textit{89.68} & \textit{93.12} & \textit{73.55} & \textit{94.85} & \textit{81.19} \\
  \midrule
  \rowcolor{gray!20}
  & \textit{BERT-base} & \textbf{64.18} & \textbf{55.46} & \textbf{58.70} & \textbf{50.12} & \textbf{64.33} & \textbf{55.52} & \textbf{84.79} & \textbf{75.18} & \textbf{78.39} & \textbf{66.21} & \textbf{84.66} & \textbf{72.75} \\
  \rowcolor{gray!20}
  & \textit{BERT-large} & \underline{63.29} & \underline{54.71} & \underline{57.90} & \underline{49.09} & \underline{62.97} & \underline{54.37} & 75.34 & 66.65 & 69.55 & 56.54 & 74.44 & 63.11 \\
  \rowcolor{gray!20}
  \multirow{-3}*{\textsc{KCS}} & \textit{RoBERTa-large} & 45.42 & 39.32 & 41.59 & 37.04 & 47.70 & 41.11 & 74.21 & 65.74 & 68.56 & 55.58 & 73.75 & 62.30 \\
  \bottomrule
  \end{tabular}
  }
  \caption{
    Main results of knowledge composition selection on HotpotQA and 2WikiMultihopQA. \textit{MAX} and \textit{Random} represent the upper and lower bounds of knowledge composition selection performance, respectively. The \textbf{Bold} and \underline{underline} mark the best and second-best results, excluding \textit{MAX}. 
  }
  \label{tab:kcs}
\end{table*}

\subsection{Baselines}

\paragraph{Knowledge Composition Selection} we compare our model against five categories of baselines to identify question-worthy sentences that encapsulate contextually relevant knowledge: \\(1) \textsc{Retrieval}: We retrieve $K$ relevant sentences in context as a knowledge composition either all at once or step by step, using only the answer or the answer concatenated previously retrieved sentences as input. \\(2) \textsc{CLS}: We concatenate the answer with each sentence in context to perform a binary classification, then select the top-$K$ sentences that exhibit the positive label as a knowledge composition. \\(3) \textsc{Sentence Graph}: We construct a sentence graph based on common entities or sentence similarity, then randomly select one sentence containing the answer as the start node and perform a random walk on this graph to select $K-1$ sentences. These $K$ sentences construct a knowledge composition. \\(4) \textsc{LLM}: We employ a zero-shot approach to prompt the large language models to generate $K$ question-worthy sentences as a knowledge composition for each sample, with the context and answer as input. \\(5) \textsc{BASE}: We randomly select $K$ sentences as a knowledge composition (\textit{Random}) or assume that all sentences selected are question-worthy (\textit{MAX}). \textit{MAX} and \textit{Random} represent the upper and lower bounds of knowledge composition selection performance, respectively.

\paragraph{Diversifying Question Generation} We evaluate the effect of diversifying question generation for data augmentation in the downstream MHQA task. Two popular LLMs, \textsc{Llama3.1} (8B) and \textsc{Qwen2.5} (7B), are employed as baseline models. Beyond supervised finetuning the baseline models on the original training data (\textit{ORI}), we compare three typical approaches for diversifying question generation: \\(1) \textit{RAST} \citep{gou-etal-2023-diversify}: This approach utilizes external knowledge to generate multiple expressions of the same question. \\(2) \textit{Composition} \citep{narayan-etal-2022-well}: This approach employs nucleus sampling to extract diverse entity chains from the context and beam search to guide question generation. \\(3) \textit{GPT-4}: We employ a zero-shot approach to prompt GPT-4 \citep{achiam2023gpt} to generate $N_q$ diverse multi-hop questions with the context and answer as input for each sample in original training data.

\begin{table*}
    \centering
    \scalebox{0.7}{
    \begin{tabular}{l|l|cccccc|cccccc}
    \toprule
     \multirow{2}*{\textbf{LLM}} & \multirow{2}*{\textbf{Diversifying Method}} & \multicolumn{6}{c}{\textbf{\textsc{HotpotQA}}} & \multicolumn{6}{c}{\textbf{\textsc{2WikiMultihopQA}}} \\
     & & \textit{EM} & \textit{P} & \textit{R} & \textit{F1} & \textit{BSc} & \textit{HSc} & \textit{EM} & \textit{P} & \textit{R} & \textit{F1} & \textit{BSc} & \textit{HSc} \\
    \midrule
    \multirow{5}*{\textsc{Llama3.1}} & \textit{ORI} & 54.00 & 66.07 & 65.29 & 64.46 & 58.74 & 69.05 & 49.00 & 58.32 &	58.54 &	57.64 &	51.36 &	62.65 \\
     & \quad$+$\textit{RAST} & \underline{55.50} & \underline{67.80} & \underline{68.33} & \underline{66.49} & 59.93 & 71.40 & 55.00 & 63.50 & 63.75 & 62.88 & 57.83 & 67.65 \\
     & \quad$+$\textit{Composition} & 55.00 & 67.17 & 67.48 & 66.09 & \underline{61.82} & \underline{72.65} & \underline{60.50} & \underline{67.45} & \underline{67.75} & \underline{66.92} & \underline{63.91} & \underline{72.45} \\
     & \quad$+$\textit{GPT4} & 49.50 & 60.81 & 61.45 & 59.75 & 52.43 & 65.55 & 55.00 & 62.62 & 63.95 &	62.51 &	58.28 &	67.60 \\
     \rowcolor{gray!20}
     & \quad$+$\textit{KCS} & \textbf{61.00} & \textbf{73.59} & \textbf{72.58} & \textbf{72.04} & \textbf{66.51} & \textbf{77.75} & \textbf{66.50} & \textbf{74.25} & \textbf{75.20} & \textbf{73.91} & \textbf{69.53} & \textbf{78.10} \\
    \midrule
    \multirow{5}*{\textsc{Qwen2.5}} & \textit{ORI}  & 39.50 & 54.90 & 51.46 & 51.36 & \underline{47.16} & \underline{60.40} & 37.50 & 44.45 &	44.02 &	43.77 &	37.74 &	52.60 \\
     & \quad$+$\textit{RAST} & 34.50 & 49.57 & 46.66 & 45.96 & 39.86 & 54.85 & 40.00 & 45.62 & 45.57 & 45.23 &	39.93 &	53.80 \\
     & \quad$+$\textit{Composition} & \underline{40.50} & \underline{55.70} & \underline{52.29} & \underline{52.02} & 46.43 & 59.70 & \underline{44.50} & \underline{49.52} & \underline{49.72} & \underline{49.27} & \underline{43.20} &	\underline{56.90} \\
     & \quad$+$\textit{GPT-4} & 32.50 & 45.93 & 42.51 & 42.33 & 35.14 & 50.75 & 41.50 & 46.27 & 47.41 & 46.27 &	39.74 &	53.90 \\
     \rowcolor{gray!20}
     & \quad$+$\textit{KCS} & \textbf{45.00} & \textbf{60.49} & \textbf{58.54} & \textbf{57.40} & \textbf{51.56} & \textbf{67.55} & \textbf{51.50} & \textbf{56.19} & \textbf{57.16} & \textbf{56.35} & \textbf{51.44} & \textbf{64.00} \\
    \bottomrule
    \end{tabular}
    }
    \caption{
      Comparison of performance for \textsc{Llama3.1} and \textsc{Qwen2.5} on \textsc{HotpotQA} and \textsc{2WikiMultihopQA}. Original represents training LLMs using the original training data. The \textbf{Bold} and \underline{underline} mark the best and second-best results.
    }
    \label{tab:mgqa}
\end{table*}

\subsection{Implementation Details}

We utilize BERT-base\footnote{\url{https://huggingface.co/google-bert/bert-base-uncased}} and MixQG-base\footnote{\url{https://huggingface.co/Salesforce/mixqg-base}} as our foundational models for knowledge composition selection and multi-hop question generation, respectively. To train the knowledge composition selection model, we preprocess the supporting facts in the training data to to obtain the (gold) knowledge composition. Specifically, for each training sample, we prioritize the document containing the answer, and within the same document, we adhere to the contextual sentence order. The sentence containing the answer is used to split the sentences of the document into two sequences that maintain the contextual order, with the sequence containing the answer positioned earlier. The model is trained with $\lambda$ of $1$ and $\rm MI$ is a cosine function. For knowledge composition selection, we employ a greedy sampling strategy for evaluation. For diversifying knowledge composition, KCS generates $N_q=5$ knowledge compositions for each sample, using a nucleus sampling strategy with top-$p=0.95$, and each composition contains $K=3$ pieces of knowledge. These obtained knowledge compositions are then used to generate $N_q$ diverse multi-hop questions. Our implementation is in PyTorch\footnote{\url{https://pytorch.org/}}, using AdamW for optimization with a learning rate of $3\times10^{-5}$ and a linear warmup ratio of $0.1$. 

For the downstream MHQA task, we focus on the distractor setting, where supporting documents include distractor documents, challenging the model to handle noise in the input. After obtaining the augmented training data by diversifying methods with $N_q=5$, we fine-tune LLMs using LoRA \citep{hu2022lora}, with a learning rate of $5\times 10^{-5}$, a cosine warmup ratio of $0.1$, a LoRA rank of $8$, and a LoRA alpha of $32$. We conduct experiments with Composition Sampling (\textit{Composition}) \citep{narayan-etal-2022-well} by ourselves, fine-tuning Pegasus\footnote{\url{https://huggingface.co/google/pegasus-large}}, and employing the nucleus sampling and beam search to obtain
diverse entity compositions and generate the most-likely multi-hop questions, respectively. 

\subsection{Main Results}

\begin{figure}[ht]
  \centering
  \includegraphics[width=\columnwidth]{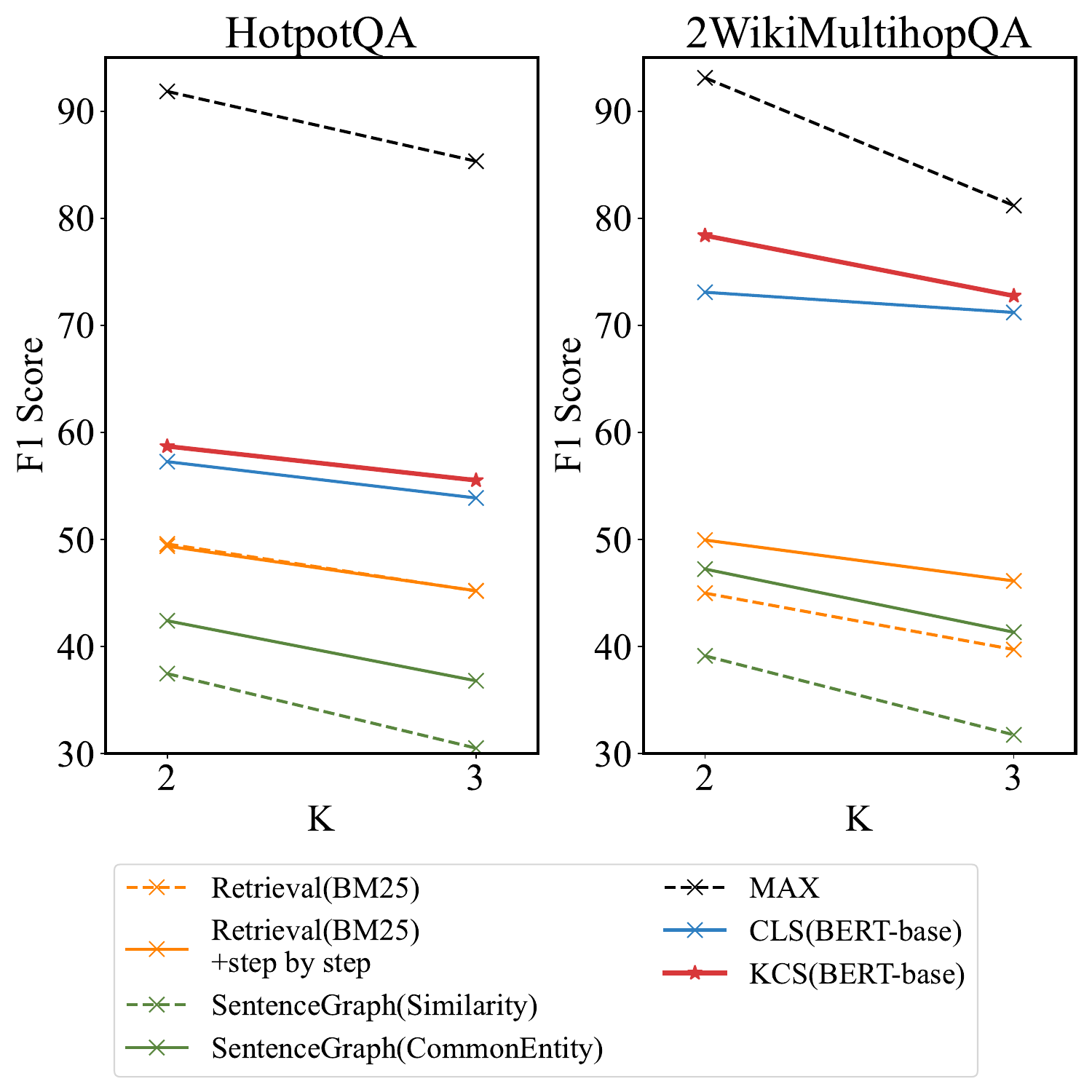}
  \caption{Accuracy comparison of competitive baselines for knowledge composition selection on HotpotQA and 2WikiMultihopQA.}
  \label{fig:selection}
\end{figure}

\paragraph{Knowledge Composition Selection} Table~\ref{tab:kcs} presents the results of knowledge composition selection on HotpotQA and 2WikiMultihopQA datasets. Each block includes a category of baselines. Our \textsc{KCS} method consistently achieves the highest scores for P, R, and F1 metrics across all datasets (HotpotQA and 2WikiMultihopQA) and knowledge composition lengths ($K=2,3$). Among the baselines, the classification (\textsc{CLS}) category yields the best results due to the high correlation between knowledge compositions and answers in datasets. Compared to \textsc{CLS} using BERT-base, our \textsc{KCS} improves the overall accuracy by $3.9\%$.  Specifically, on HotpotQA, KCS achieves $58.70$ (F1@2) and $55.52$ (F1@3) with the knowledge composition length $K=2,3$, respectively. These scores represent $63.90\%$ and $65.06\%$ of the upper bound (\textit{MAX} of \textsc{BASE}), and outperform the most competitive \textsc{CLS} baseline by about $1.5\%$ and $3\%$. On 2WikiMultihopQA, \textsc{KCS} demonstrates even stronger performance, with $84.18\%$ and $89.60\%$ of the upper bound and about $7.2\%$ and $2.1\%$ outperform to \textsc{CLS}. Surprisingly, the performance of \textsc{KCS} is further improved when the length of knowledge composition is increased from $2$ to $3$ during inference. This phenomenon indicates that \textsc{KCS} has higher accuracy in predicting longer knowledge compositions and has effectively learned the potential knowledge coherence. Visual comparison of the accuracy is shown in Figure~\ref{fig:selection}. Our ablation study further investigates the significant impact of knowledge composition length and order on model performance during training. Despite lack of annotated sequential relationships between knowledge and answers in training data, results indicate that \textsc{KCS} efficiently achieves cost-effectively and high-performance knowledge composition selection.

\paragraph{Diversifying Question Generation} 

The MHQA performance of \textsc{Llama3.1} and \textsc{Qwen2.5} finetuned on data augmented by recent diversifying methods is shown in Table~\ref{tab:mgqa}. The results indicate that using KCS for data augmentation achieves consistent improvements on both HotpotQA and 2WikiMultihopQA datasets, which illustrate that sampled knowledge compositions are meaningful to a certain extent. And we provide a detailed case study in Figure~\ref{fig:case}. To further illustrate that the sampled knowledge compositions are logically coherent, we conduct a LLM evaluation and an actual human analysis in  Appendix~\ref{appendix:logic}. Collectively,  these results substantiate that the pieces of knowledge in sampled knowledge compositions can be logically combined to generate valid multi-hop questions.

To better understand the impact of question diversity on downstream MHQA performance, we assess the consistency and diversity of generated questions in Table~\ref{tab:dqg}. The results indicate that \textit{KCS} achieves high diversity, with consistency limited by the vanilla MHQG method. As discussed in Section~\ref{sec:mhqg}, advanced multi-hop question generation methods can address this issue. \textit{RAST} and \textit{Composition} exhibit higher consistency due to repeated high-consistency questions, which negatively impacts diversity. Although \textit{GPT-4} achieves high diversity and consistency, it incurs prohibitive costs. We attribute the helpful improvement on MHQA performance to the effectiveness of KCS in balancing high question diversity with minimal noise.

\begin{table}[h]
  \centering
  \scalebox{0.73}{
  \begin{tabular}{l|cccc}
  \toprule
  \multirow{2}*{\textbf{Metrics}} & \multicolumn{4}{c}{\textbf{\textbf{Diversifying Methods}}} \\
   & \textit{Composition} & \textit{RAST} & \textit{GPT-4} & \textit{KCS} \\
  \midrule
  \textit{Pairwise-BLEU} ($\downarrow$) & 89.5 & 71.4 & \textbf{15.0} & \underline{68.1} \\
  \textit{LLM-Diversity} ($\uparrow$) & 29.0 & 38.6 & \textbf{79.8} & \underline{46.2} \\
  \textit{LLM-Consistency} ($\uparrow$) & \underline{74.8} & 69.6 & \textbf{91.8} & 68.0 \\
  \bottomrule
  \end{tabular}
  }
  \caption{Diversity and consistency of different diversifying methods on HotpotQA. The \textbf{Bold} and \underline{underline} mark the best and second-best results of each row.}
  \label{tab:dqg}
\end{table}

\subsection{Ablation Study}

We investigate the impact of various factors on the performance of the KCS framework.

\paragraph{Decoder-Only vs. Encoder-Decoder Architectures} We compare decoder-only and encoder-decoder architectures for knowledge composition selection. We remove the encoder of $\mathcal{M}_{\rm seq}$, and retrain the $\mathcal{M}_{\rm enc}$ and the decoder of $\mathcal{M}_{\rm seq}$ to form the decoder-only architecture (\textit{Decoder}). The \textit{Decoder} autoregressively selects the next sentence from the long context, using the answer and previously selected sentences as input. As shown in Table~\ref{tab:ab}, the decoder-only model (\textit{Decoder}, the top block) achieves $47.74$ (F1@1), $55.94$ (F1@2) and $52.89$ (F1@3), indicating the critical role of the decoder in KCS performance. Incorporating the encoder and a classification objective (\textit{Encoder-Decoder}, the bottom block) further improves performance by $1.26$, $2.76$ and $2.63$ points, respectively. The \textit{-decoder} setup in the bottom block, which uses encoder classification probabilities for knowledge composition selection, demonstrates a significant drop by $16.12$, $15.08$ and $11.62$, also highlighting the importance of the decoder.

\paragraph{Knowledge Composition Order and Length in Training} We assess the \textit{Decoder} performance across various sentence arrangements within knowledge compositions: (1) \textit{original}: Original sequence of sentences in the dataset. (2) \textit{shuffle}: Randomized order of sentences at each training epoch. (3) \textit{sorted}: Sentences arranged according to their contextual order. (4) \textit{cluster}: Consecutive sentences grouped together, prioritizing clusters that contain answers. (5) \textit{cropping}: A maximum of 2 sentences per composition. (6) \textit{document}: Documents containing answers are ranked highly and then reordered based on their contextual order in the document to create a (gold) knowledge composition. As shown in Table~\ref{tab:ab}, the \textit{document} arrangement proves most efficient, with different orders causing a maximum performance drop by $17.2$ (F1@1), $5.56$ (F1@2) and $5.29$ (F1@3). A cropping length of 2 results in a performance drop by $4.49$ (F1@1), $4.46$ (F1@2) and $3.43$ (F1@3). These results indicate that both order and length of knowledge composition during training significantly influence KCS performance.

\paragraph{Pre-Encoder vs. Post-Encoder  Concatenation} We compare accuracy using answer representation concatenated with context representations before (\textit{pre}) and after (\textit{post}) encoder layers in the \textit{Encoder-Decoder} architecture. Results in Table~\ref{tab:ab} indicate that \textit{post} outperforms \textit{pre} by $0.2$ (F1@1), $0.96$ (F1@2) and $0.97$ (F1@3) points, respectively.

\paragraph{Hyper-parameters of Architecture and Training} Based on the best setup (\textit{Encoder-Decoder} with \textit{post}), we further explore the effects of expanding transformer layers and attention heads on model performance (\textit{4l8h}). Our experiments show that expanding transformer layers ($2\rightarrow 4$) and attention heads ($4\rightarrow 8$) does not yield improvements, resulting in drops by $1.77$ (F1@1), $2.47$ (F1@2) and $2.27$ (F1@3) points. Additionally, adjusting the $\lambda$ parameter from $1$ to $0.5$ (\textit{0.5$\lambda$}) results in performance drops of $0.76$ (F1@1), $0.99$ (F1@2) and $1.05$ (F1@3) points, highlighting the importance of the probabilistic contrastive loss.

\paragraph{w/o Integration of Knowledge Classification Probabilities} The \textit{cls\&gen} setup does not integrate the knowledge classification probabilities into the decoder, i.e. hidden states $H$ are used as both the key and the value states. Results in Table~\ref{tab:ab} show that \textit{cls\&gen} results in performance drops of $0.08$ (F1@1), $0.77$ (F1@2) and $0.82$ (F1@3) points, indicating that the integration of knowledge classification probabilities is more effective.

\begin{table}[h]
  \centering
  \scalebox{0.9}{
  \begin{tabular}{l|ccc}
  \toprule
  \textbf{Model} & \textit{F1@1} & \textit{F1@2} & \textit{F1@3} \\
  \midrule
  \textit{Decoder}      & & & \\
  \textit{\quad$+$original} & 30.54 &	50.38 & 47.60 \\
  \textit{\quad$+$shuffle}   & 42.74 & 51.48 & 50.05 \\
  \textit{\quad$+$sorted}    & 42.72 & 51.66 & 49.90 \\
  \textit{\quad$+$cluster}   & 47.21 & 54.63 & 51.68 \\
  \textit{\quad\quad$+$cropping}    & 42.79 & 51.48 & 49.46 \\
  \rowcolor{gray!20}
  \textit{\quad$+$document} & 47.74 & 55.94 & 52.89 \\
  \midrule
  \textit{Encoder-Decoder} & & & \\
  \textit{\quad$+$pre} & 48.80 & 57.74 & 54.55 \\
  \rowcolor{gray!20}
  \textit{\quad$+$after}  & \textbf{49.00} & \textbf{58.70} & \textbf{55.52} \\
  \textit{\quad\quad$+$4l8h}  & 47.23 & 56.23 & 53.25 \\
  \textit{\quad\quad$+$cls\&gen}  & \underline{48.92} & \underline{57.93} & \underline{54.70} \\
  \textit{\quad\quad$+$0.5$\lambda$}  & 48.24 & 57.71 & 54.47 \\
  \textit{\quad\quad$-$decoder}  & 32.88 & 43.62 & 43.90 \\
  \bottomrule
  \end{tabular}
  }
  \caption{Results of ablation study for the KCS. The \textcolor{gray}{gray} row indicate the best for each block.}
  \label{tab:ab}
\end{table}

\subsection{Case Study}

To further analyze the performance of KCS on diversifying question generation, we present a case study in Appendix~\ref{appendix:case}. As illustrated in Figure~\ref{fig:case}, our KCS method diversifies subsequent knowledge based on the given answer, context, and previously selected knowledge. As the length of knowledge compositions increases, the generated multi-hop questions become more specific and complex. Compared to other baselines for diversifying QG, questions generated by KCS exhibit greater diversity at the knowledge level. Although the question generated based on the knowledge composition \texttt{[10, 43]} is inconsistent with the answer ``Loveless'', we attribute this to the absence of robust and advanced multi-hop question generation methods rather than knowledge selection issues, since we can easily utilize ``My Bloody Valentine'' as bridging content in sentences \texttt{10} and \texttt{43} to formulate a multi-hop question.

\section{Conclusion}

This paper introduces KCS, a novel framework designed to expand the diversity of generated multi-hop questions by sampling varied knowledge compositions within a given long context. Unlike prior methods that rely on structured graphs or fixed knowledge compositions, KCS leverages unstructured text to discern the potential knowledge coherence, making it more flexible and scalable. To mitigate the risk of degeneration, we propose sentence-level conditional prediction and a probabilistic contrastive loss to learn the potential knowledge coherence. To balance the accuracy and diversity, we employ a stochastic decoding strategy that truncates the unreliable tail of the probability distribution and samples the next knowledge piece from a dynamic nucleus. Comprehensive experiments show that KCS improves the overall accuracy of knowledge composition selection and its application for data augmentation enhances downstream performance.

\section*{Limitations}

Our work currently exists several limitations and future directions: (1) KCS still has room for optimization. Explore different ways to calculate mutual information (such as cosine or inner product), various model architectures\footnote{For example, to use a MoE LoRA architecture \citep{tang2025graphmoe}.}, advanced multi-hop question generation methods and the availability of high-quality annotated data may help improve KCS; (2) When use for data augmentation, investigating the performance of KCS on domain-specific data can help mitigate data sparsity challenges in domains with low resources; (3) KCS demonstrates the potential of Transformer model in high-level representation prediction and the advantages of probabilistic contrastive loss, which may inspire other similar works; (4) Not any diverse data is beneficial to downstream task improvement, and figuring out what kind of data is valuable for downstream tasks is also a direction worth studying.

\section*{Acknowledgments}

We gratefully acknowledge the support of the National Natural Science Foundation of China (Grant No. 62350710797), the Key Research and Development Program of Heilongjiang Province, China [2024ZX01A07], and the Science and Technology Innovation Award of Heilongjiang Province, China [JD2023GJ01].

\bibliography{custom}

\cleardoublepage
\appendix 

\section{Dataset statistics}
\label{appendix:statistics}

The detailed statistics of both HotpotQA and 2WikiMultihopQA datasets are provided in Table~\ref{tab:statistic}.

\begin{table}[ht]
  \centering
  \scalebox{0.75}{
  \begin{tabular}{l|cc}
  \toprule
  \multirow{2}*{\textbf{Dataset}} & \textbf{\textsc{HotpotQA}} & \textbf{\textsc{2WikiMultihopQA}} \\
  & \textit{Min/Mean/Max} & \textit{Min/Mean/Max} \\
  \midrule
  Answer Len.   & 3/5.2/171      & 3/5.1/28        \\
  Question Len. & 6/24.7/143     & 8/17.6/58       \\
  Sentence Len. & 2/30.8/590     & 3/25.0/375      \\
  Context Len.  & 50/1273.9/3764 & 125/825.6/5967  \\
  KC Len.     & 9/83.6/424     & 16/72.0/314     \\
  Context Num.  & 2/41.2/147     & 10/33.0/209     \\
  KC Num.     & 2/2.4/9        & 2/2.2/5         \\
  \midrule
  & \textit{Train/Dev/Test} & \textit{Train/Dev/Test}\\
  Sample Num.   & 84487/479/6947 & 109589/306/11281 \\
  \bottomrule
  \end{tabular}
  }
  \caption{Dataset statistics of HotpotQA and 2WikiMultihopQA, where Len. and Num. denote the number of tokens and the number of sentences, respectively. KC denotes the knowledge composition.}
  \label{tab:statistic}
\end{table}

\section{Detailed Example}
\label{appendix:example}

\begin{figure}[h]
  \centering
  \includegraphics[width=0.95\columnwidth]{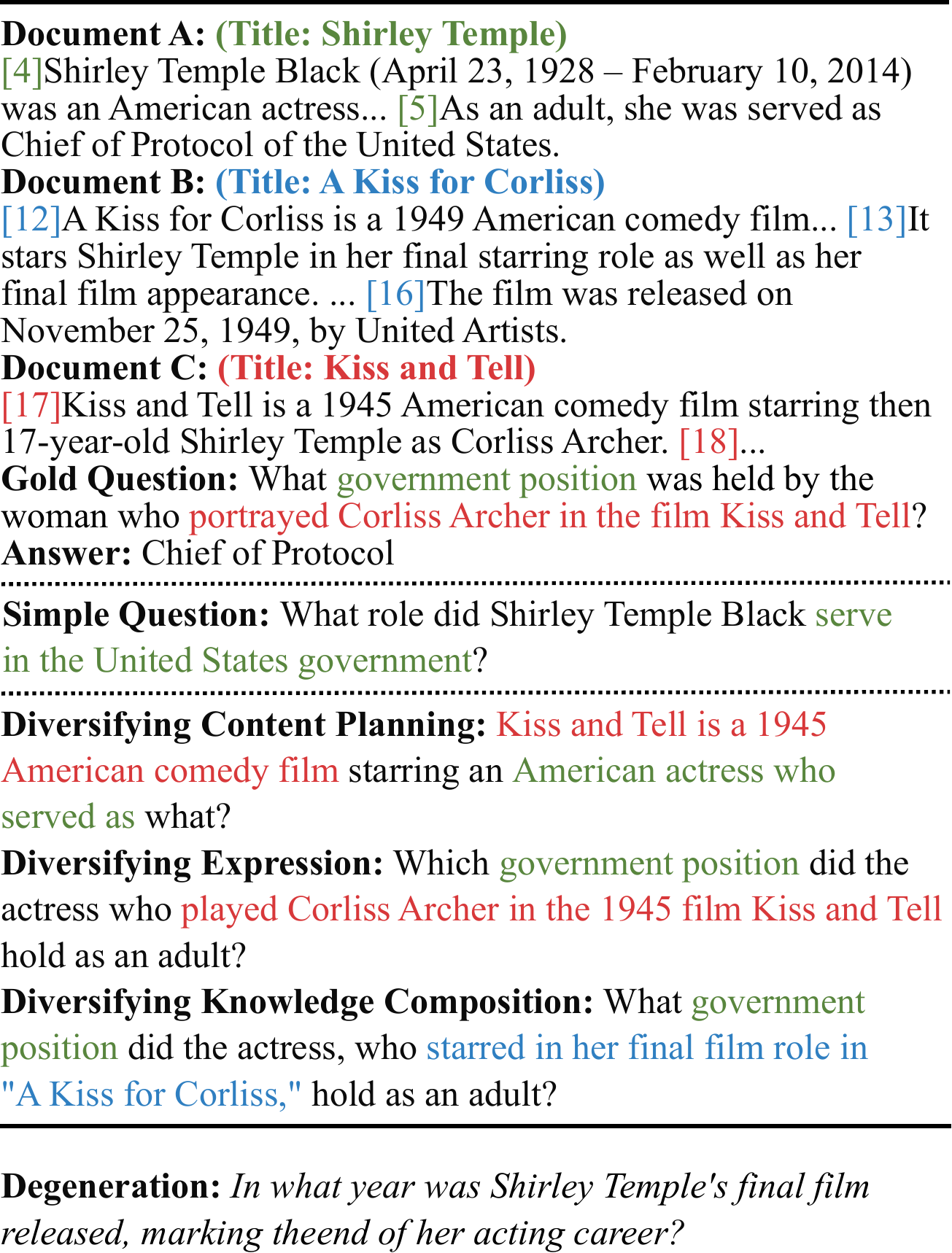}
  \caption{A detailed example. Colored numbers \textcolor{green}{$\bullet$}\textcolor{blue}{$\bullet$}\textcolor{red}{$\bullet$} indicating sentences from different documents.}
  \label{fig:_example}
\end{figure}

The detailed example is shown in Figure~\ref{fig:_example}, which is algined with the example in Figure~\ref{fig:example}.

\section{Case Study}
\label{appendix:case}

\begin{figure}[!ht]
  \centering
  \includegraphics[width=\columnwidth]{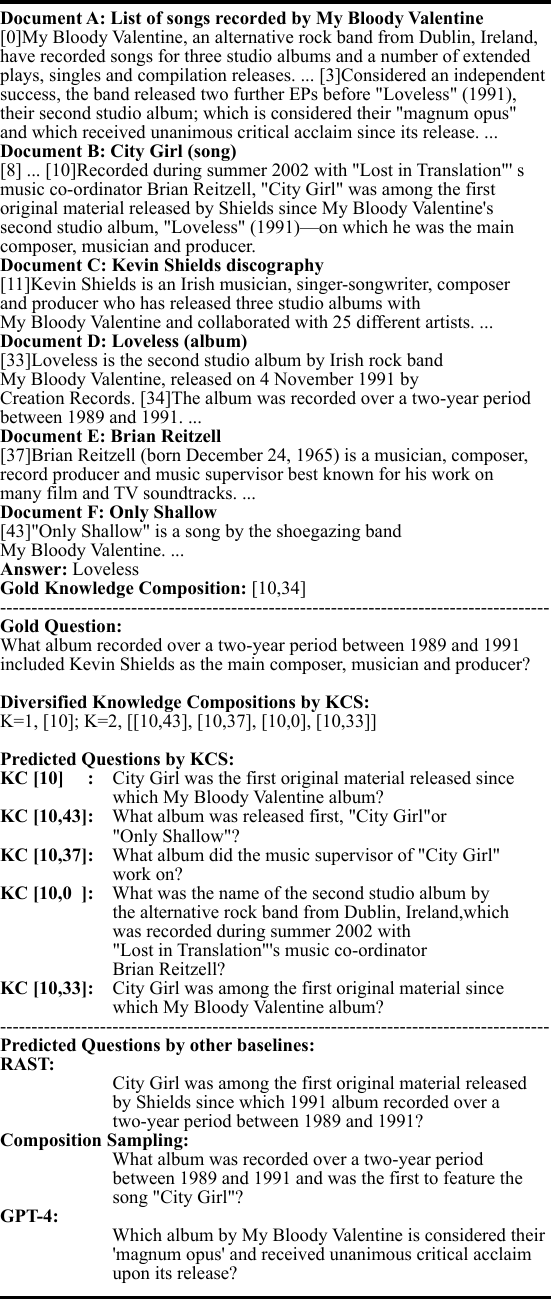}
  \caption{Case study.}
  \label{fig:case}
\end{figure}

To further analyze the impact of diverse knowledge compositions obtained through our Knowledge Composition Sampling (KCS) method on diversifying multi-hop question generation, we present a case study in Figure~\ref{fig:case}.

\section{Logical Coherence Study}
\label{appendix:logic}

To demonstrate the logical coherence of the sampled knowledge compositions, we conducted evaluations using both a large language model (LLM) and human analysis.

Initially, we employ GPT-4 to assess whether the sampled compositions can be logically combined to form valid multi-hop questions. Specifically, we sample 200 examples from the original training data of the downstream MHQA task. We compare the LLM evaluation scores of KCS, which involves sampling five compositions per example, against the LLM evaluation scores of the ground truth, which includes one annotated meaningful knowledge composition per example. As illustrated in Table~\ref{tab:logic}, the LLM evaluation scores for KCS closely approximate those of the ground truth, while significantly enhancing scale and diversity. Quantitatively, the LLM evaluation scores for KCS achieve 91.55\% and 93.32\% of the ground truth scores on HotpotQA and 2WikiMultihopQA, respectively. This finding confirms that the knowledge compositions selected by KCS can be logically combined to form valid multi-hop questions.

\begin{table}[h]
  \centering
  \scalebox{0.73}{
  \begin{tabular}{l|cc}
  \toprule
  \textbf{Data(Example num)} & {\textsc{\textbf{HotpotQA}}} & {\textsc{\textbf{2WikiMultihopQA}}} \\
  \midrule
  \textit{Ground Truth}(200) & 80.95 & 75.00 \\
  \textit{KCS}(1000) & 74.11 & 69.99 \\
  \bottomrule
  \end{tabular}
  }
  \caption{LLM evaluation scores for original ground truth and KCS-diversified knowledge compositions on HotpotQA and 2WikiMultihopQA.}
  \label{tab:logic}
\end{table}

\begin{figure}[!ht]
\centering
\includegraphics[width=\columnwidth]{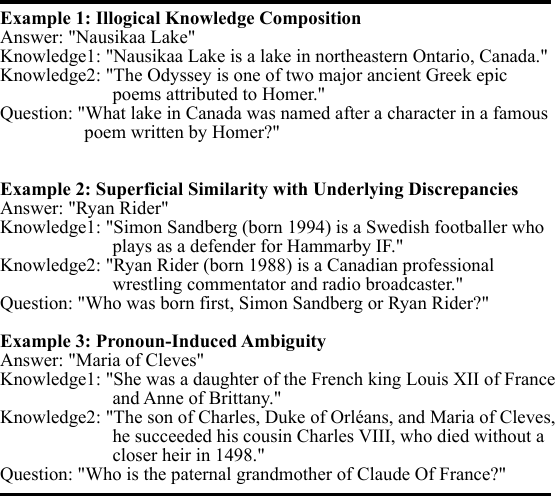}
\caption{Examples from the logical coherence study.}
\label{fig:logic}
\end{figure}

Since the LLM evaluation has a 20\% to 25\% deviation from the human-annotated ground truth, we conduct a manual review and in-depth analysis of these samples. Some cases are contentious, which does not invalidate the efficacy of LLM-Eval, as human evaluators may also produce divergent judgments. Other cases contain errors in human annotations, predominantly characterized by redundant knowledge components. Overall, these samples can be categorized into three distinct groups: (1) Illogical Knowledge Composition: Some combinations of knowledge components are redundant and lack coherent logical relationships; (2) Superficial Similarity with Underlying Discrepancies: Some knowledge components appear similar in form but represent fundamentally distinct facts, leading to ambiguity in their relationship (even for human evaluators); (3) Pronoun-Induced Ambiguity: Some knowledge components contain excessive pronominal references, resulting in semantic ambiguity that complicates relationship identification (even for human evaluators). Examples are provided in Figure~\ref{fig:logic}.

The results of logical coherence study demonstrate that the selected knowledge components in KCS can be logically combined to generate valid multi-hop questions. The improved performance by KCS on downstream MHQA task also provides additional validation for the logical coherence in sampled knowledge compositions.

\section{Algorithm of Diversifying Multi-hop QG}
\label{appendix:algorithm}

The diversifying multi-hop QG algorithm is detailed in Algorithm~\ref{alg:pipeline}, which is discussed in Section~\ref{sec:overview}.

\SetAlCapFnt{\small}
\SetAlCapNameFnt{\small}
\SetAlgoNlRelativeSize{-1}

\begin{algorithm}
\small

\caption{Diversify Multi-hop QG}
\label{alg:pipeline}

\KwIn{Context $D$, answer $a$, and $K,N_q,p$,}
\KwOut{Diversified multi-hop questions $Q$}

$Q \leftarrow \emptyset$\;

\For{$i \leftarrow 1$ \KwTo $N_q$}{
    $c_i \leftarrow \emptyset$\;
    \For{$k \leftarrow 1$ \KwTo $K$}{
        $e_k \leftarrow$ generate latent prediction representation with input $(D, a, c_i)$\;
        $p_{\rm new} \leftarrow$ rescaled $p(s|e_k)$ to a new distribution with $p$\;
        $s \leftarrow$ randomly sample a sentence from $D$ based on $p_{\rm new}$\;
        $c_i \leftarrow c_i \cup \{s\}$\;
    }
    $q \leftarrow$ generate question from $(c_i, a)$\;
    $Q\leftarrow Q \cup \{q_i\}$\;
}

\Return $Q$\;

\end{algorithm}

\end{document}